\documentclass{article}
\usepackage{spconf,amsmath,graphicx}
\usepackage{multirow, hhline, stmaryrd}

\title{Unsupervised Adaptation with Domain Separation Networks for Robust Speech
Recognition}
\name{Zhong Meng$^{1,2}$\sthanks{Zhong Meng performed the work while he was a research intern
	at Microsoft AI and Research, Redmond, WA.}, Zhuo Chen$^{1}$, Vadim
	Mazalov$^{1}$, Jinyu Li$^{1}$, Yifan Gong$^{1}$
}
\address{$^{1}$ Microsoft AI and Research, Redmond, WA
\\ $^{2}$ Georgia Institute of Technology, Atlanta, GA
} 
\begin{document}
%
\maketitle
\begin{abstract}
Unsupervised domain adaptation of speech signal aims at adapting a well-trained
source-domain acoustic model to the unlabeled data from target
domain. This can be achieved by adversarial training of deep neural network
(DNN) acoustic models to learn an intermediate deep representation that is both
senone-discriminative and domain-invariant. Specifically, the DNN is trained to
jointly optimize the primary task of senone classification and the secondary task of
domain classification with adversarial objective functions. In this work, instead of
only focusing on learning a domain-invariant feature (i.e. the shared component
between domains), we also characterize the difference between the source and target
domain distributions by explicitly modeling the private component of each domain
through a private component extractor DNN. The private component is trained to be orthogonal with the
shared component and thus implicitly increases the degree of domain-invariance of the
shared component. A reconstructor DNN is used to reconstruct the original speech feature
from the private and shared components as a regularization. This domain separation
framework is applied to the unsupervised environment adaptation task and achieved
11.08\% relative WER reduction from the gradient reversal layer training, a
representative adversarial training method, for automatic speech recognition on CHiME-3 dataset.
\end{abstract}
\section{Introduction}
\label{sec:intro}
In recent years, advances in deep learning have led to remarkable performance boost in automatic speech recognition (ASR)  \cite{seide2011conversational, sainath2011making, jaitly2012application, DNN4ASR-hinton2012, deng2013recent, yu2017recent}. 
However, ASR systems still suffer from large performance degradation when acoustic mismatch exists between the training and test conditions \cite{Li14overview, Li15robust}.
Many factors contribute to the mismatch, such as variation in environment noises, channels and speaker characteristics. 
Domain adaptation is an effective way to address this limitation, in which
the acoustic model parameters or input features are adjusted to compensate for the mismatch.

One difficulty with domain adaptation is that available data from the target domain is usually limited, in which case the acoustic model can be easily overfitted.
To address this issue, regularization-based approaches are proposed in
\cite{kld_yu, map_huang, l2_liao, huang2015rapid} to regularize the neuron output distributions or the model parameters. 
In \cite{lhn, feature_seide}, transformation-based approaches are introduced to reduce the number of learnable parameters. In \cite{svd_xue_1,svd_xue_2, svd_zhao}, the trainable parameters are further reduced by singular value decomposition of weight matrices of a neural network. 
Although these methods utilize the limited data from the target domain, they still require labelling for the adaptation data and can only be used in supervised adaptation. 

Unsupervised domain adaptation is necessary when human labelling of the target domain data is unavailable. It has become an important topic with the rapid increase of the amount of untranscribed speech data for which the human annotation is expensive. Pawel et al. proposed to learn the contribution of hidden units by additional amplitude parameters \cite{lhuc_pawel_1} and differential pooling \cite{lhuc_pawel_2}. Recently, Wang et al. proposed to adjust the linear
transformation learned by batch normalized acoustic model in \cite{bn_wang}. Although these methods lead to increased performance in the ASR task when no labels are available for the adaptation data, they still rely on the senone (tri-phone state) alignments against the unlabeled adaptation data through first pass decoding. The first pass decoding result is unreliable when the mismatch between the training and test conditions is significant. It is also time-consuming and can be hardly applied to huge amount of adaptation data. There are even situations when decoding adaptation data is not allowed because of the privacy agreement signed with the speakers. These methods depending on the first pass decoding of the unlabeled adaptation data is sometimes called ``semi-supervised'' adaptation in
literature.

The goal of our study is to achieve \emph{purely} unsupervised domain adaptation
\emph{without} any exposure to the labels or the decoding results of the adaptation
data in the target domain. In \cite{li2017large} we show that the source-domain model can be effectively adapted without any transcription by using teacher-student (T/S) learning \cite{li2014learning}, in which the posterior probabilities generated by the source-domain model can be used in lieu of labels to train the target-domain model. However, T/S learning relies on the availability of parallel unlabeled data which can be usually simulated. However, if parallel data is not available, we cannot use T/S learning for model adaptation. In this study, we are exploring the solution to domain adaptation without parallel data and without transcription. Recently, adversarial training has become a very hot topic
in deep learning because of its great success in estimating generative models
\cite{gan}. It was first applied to the area of unsupervised domain adaptation by
Ganin et al. in \cite{grl_ganin} in a form of multi-task learning. In their work, the
unsupervised adaptation is achieved by learning deep intermediate representations
that are both discriminative for the main task (image classification) on the source
domain and invariant with respect to mismatch between source and target domains. The
domain invariance is achieved by the adversarial training of the domain
classification objective functions.  This can be easily implemented by augmenting any
feed-forward models with a few standard layers and a \emph{gradient reversal layer
(GRL)}. This GRL approach has been applied to acoustic
models for unsupervised adaptation in \cite{grl_sun} and for increasing noise robustness in
\cite{grl_shinohara, grl_serdyuk}. Improved ASR performance is achieved in both
scenarios.

However, the GRL method focuses only on learning a domain-invariant representation, ignoring the unique
characteristics of each domain, which could also be informative. 
Inspired by this,
Bousmailis et al. \cite{dsn} proposed the \emph{domain separation networks (DSNs)} to
separate the deep representation of each training sample into two parts: one private
component that is unique to its domain and one shared component that is invariant to
the domain shift.  
In this work, we propose to apply DSN for unsupervised domain adaptation on a DNN-hidden Markov model (HMM) acoustic model, aiming to increase the noise robustness in speech recognition. 
In the proposed framework, the shared component is learned to be both senone-discriminative and domain-invariant through adversarial multi-task training of a shared component extractor and a domain classifier. 
The private component is
trained to be orthogonal with the shared component to implicitly increase the degree
of domain-invariance of the shared component. A reconstructor DNN is used to
reconstruct the original speech feature from the private and shared components, serving for 
regularization.  The proposed method achieves $11.08\%$ relative WER improvement over
the GRL training approach for robust ASR on the CHiME-3 dataset.


\section{Domain Separation Networks}
\label{sec:dsn}
In the \emph{purely} unsupervised domain adaptation task, we only have access to a sequence of speech frames
$X^s=\{x^s_{1}, \ldots, x^s_{N_s}\}$ from the source domain distribution, a sequence
of senone labels $Y^s=\{y^s_{1},\ldots, y^s_{N_s}\}$
aligned with source data $X^s$ and a sequence of speech frames $X^t=\{x^t_{1},
\ldots, x^t_{N_t}\}$ from a target domain distribution. Senone labels or other types of transcription are \emph{not} available for the target speech sequence $X^t$.

When applying domain separation networks (DSNs) to the unsupervised adaptation task, our
goal is to learn the shared (or common) component extractor DNN $M_c$ that maps an
input speech frame $x^s$ from source domain or $x^t$ from target domain to a
\emph{domain-invariant} shared component $f^s_c$ or $f^t_c$ respectively. At the same time, learn a senone
classifier DNN $M_y$ that maps the shared component $f^s_c$ from the source domain to the correct
senone label $y^s$. 

To achieve this, we first perform adversarial training of the domain classifier DNN
$M_d$ that maps the shared component $f^s_c$ or $f^t_c$ to its domain label $d^s$ or
$d^t$, while simultaneously minimizing the senone classificaton loss of $M_y$ given
shared component $f^s_c$ from the source domain to ensure the
\emph{senone-dicriminativeness} of $f^s_c$.

For the source or the target domain, we extract the source or the target private
component $f^s_p$ or $f^t_p$ that is unique to the source or the target domain
through a source or a target private component extractor $M^s_p$ or $M^t_p$. The
shared and private components of the same domain are trained to be orthogonal to each
other to further enhance the degree of domain-invariance of the shared components. The
extracted shared and private components of each speech frame are concatenated and
fed as the input of a reconstructor $M_r$ to reconstruct the input speech
frame $x^s$ or $x^t$.

\begin{figure*}[!t]
    \centering
    \includegraphics[width=13cm]{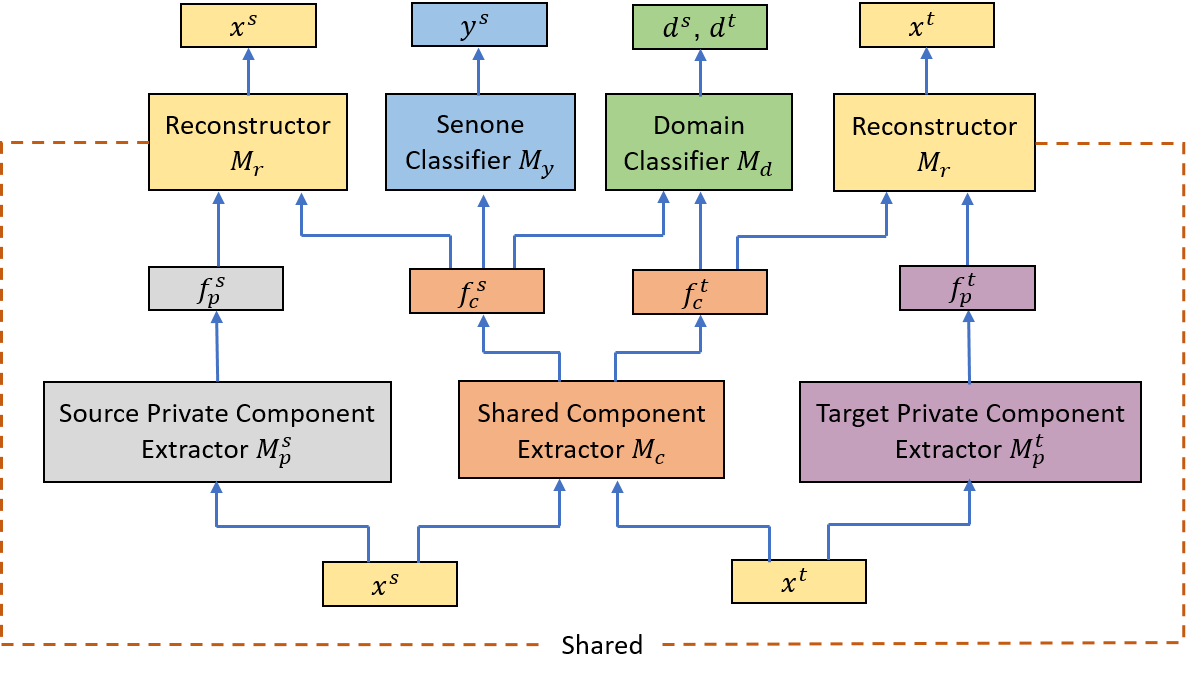}
     \vspace{-0.4cm}
    \caption{\small The architecture of domain separation networks.}
    \label{fig:dsn}
\end{figure*}

The architecture of DSN is shown in Fig. \ref{fig:dsn}, in which all the
sub-networks are jointly optimized using SGD. The optimized shared component
extractor $M^c$ and senone classifier $M_y$ form the adapted acoustic model for
subsequent robust speech recognition. 

\subsection{Deep Neural Networks Acoustic Model}
\label{sec:dnn_am}
The shared component extractor $M_c$ and senone predictor of the DSN are initialized
from an DNN-HMM acoustic model. The DNN-HMM acoustic model is trained with
labeled speech data $(X^s, Y^s)$ from the source domain. The senone-level alignment
$Y_s$ is generated by a well-trained GMM-HMM system. 

Each output unit of the DNN acoustic model corresponds to one of the senones in the set
$\mathcal{Q}$. The output unit for senone $q \in \mathcal{Q}$ is the posterior probability
$p(q|x^s_{n})$ obtained by a softmax function.



\subsection{Shared Component Extraction with Adversarial Training}
\label{sec:share}
The well-trained acoustic model DNN in Section \ref{sec:dnn_am} can be decomposed into two parts: a share
component extractor $M_c$ with parameters $\theta_c$ and a senone classifier $M_y$
with parameters $\theta_y$. An input speech frame from source domain $x^s$ is first
mapped by the $M_c$ to a K-dimensional shared component $f^s_c\in \mathcal{R}^K$.
$f^s_c$ is then mapped to the senone label posteriors  by a senone classifier $M_y$ with parameters
$\theta_y$ as follows. 

\begin{align}
	M_y(f^s_c) = M_y(M_c(x^s_i)) = p(\hat{y}^s_n = q | x^s_i; \theta_c,
\theta_y)
	\label{eqn:senone_classify}
\end{align}
where $\hat{y}^s_i$ denotes the predicted senone label for source
frame $x^s_i$ and $q \in \mathcal{Q}$.

The domain classifier DNN $M_d$ with parameters $\theta_d$ takes the shared component from source domain $f^s_c$
or target domain $f^t_c$ as the input to predict the two-dimensional domain label
posteriors as follows (the 1st and 2nd output units stand for the source and target
domains respectively). 
\begin{align}
	M_d(M_c(x^s_i)) & = p(\hat{d}^s_i = a | x^s_i; \theta_c, \theta_d), \; a \in
	\{1, 2\} \\
	M_d(M_c(x^t_j)) & = p(\hat{d}^t_j = a | x^t_j; \theta_c, \theta_d), \; a \in \{1, 2\}
	\label{eqn:domain_classify}
\end{align}
where $\hat{d}^s_i$ and $\hat{d}^t_j$ denote the predicted domain labels for the source
frame $x^s_i$ and the target frame $x^t_j$ respectively. 


In order to adapt the source domain acoustic model (i.e., $M_c$ and
$M_y$) to the \emph{unlabeled} data from target domain, we want to make the
distribution of the source domain shared component $P(f^s_c) = P(M_c(x^s))$ as close to
that of the target domain $P(f^t_c) = P(M_c(x^t))$ as possible. In
other words, we want to make the shared component domain-invariant. This
can be realized by adversarial training, in which we adjust the parameters $\theta_c$ of
shared component extractor to \emph{maximize} the loss of the domain classifier
$\mathcal{L}^c_{\text{domain}}(\theta_c)$ below while adjusting
the parameters $\theta_d$ to \emph{minimize} the loss of the domain classifier
$\mathcal{L}^{d}_{\text{domain}}(\theta_d)$ below. 

\begin{align}
	&\mathcal{L}^d_{\text{domain}}(\theta_d) = - \sum_{i}^{N_s} \log
	p(\hat{d}^s_i = 1 | x^s_i;
	\theta_d)\nonumber \\
	& \qquad \qquad \qquad \qquad - \sum_{j}^{N_t} \log p(\hat{d}^t_j = 2 | x^t_j; \theta_d)
	\label{eqn:domain_loss}  \\
	& \mathcal{L}^{c}_{\text{domain}}(\theta_c) = - \sum_{i}^{N_s} \log
	p(\hat{d}^s_i = 1 | x^s_i;
	\theta_c)\nonumber \\
	& \qquad \qquad \qquad \qquad - \sum_{j}^{N_t} \log p(\hat{d}^t_j = 2 | x^t_j;\theta_c)
	\label{eqn:domain_loss_adv}
\end{align}
This minimax competition will first increase the capability of both the shared
component extractor and the domain classifier and will eventually converge to the point where
the shared component extractor generates extremely confusing representations that domain
classifier is unable to distinguish (i.e., domain-invariant).  

Simultaneously, we minimize the loss of the senone classifier below to ensure
the domain-invariant shared component $f^s_c$ is also discriminative to senones.

\begin{align}
	\mathcal{L}_{\text{senone}}(\theta_c, \theta_{y}) = - \sum_{i}^{N_s} \log p(y^s_i |
	x^s_i;\theta_y, \theta_c)
	\label{eqn:loss_senone}
\end{align}

Since the adversarial training of the domain classifier $M_d$ and shared component extractor $M_c$ has made the distribution of the target domain shared-component $f^t_c$ as close to that of $f^s_c$ as possible, the $f^t_c$ is also senone-discriminative and will lead to minimized senone classification error given optimized $M_y$. Because of the domain-invariant property, good adaptation performance can be achieved when the target domain data goes through the network. 

\subsection{Private Components Extraction}
\label{sec:private}
To further increase the degree of domain-invariance of the shared components, we
explicitly model the private component that is unique to each domain by a private
component extractor DNN $M_p$ parameterized by $\theta_p$. $M_p^s$ and $M^t_p$ map
the source frame $x^s$ and the target frame $x^t$ to hidden representations
$f^s_p = M^s_p(x^s)$ and $f^t_p = M^t_p(x^t)$ which are the
private components of the source and target domains respectively. The private
component for each domain is trained to be orthogonal to the shared component by minimizing the difference loss below.
\begin{align}
	& \mathcal{L}_{\text{diff}}(\theta_c, \theta^s_p, \theta^t_p) \nonumber \\
	& = \left|\left| \sum^{N_s}_{i}
	M_c(x^s_i)M^s_p(x^s_i)^{\top}\right|\right|^2_F + \left|\left|\sum^{N_t}_{j}
	M_c(x^t_j)M^t_p(x^t_j)^{\top}\right|\right|^2_F
	\label{eqn:diff_loss}
\end{align}
where $||\cdot||^2_F$ is the squared Frobenius norm. All the vectors are assumed to be
column-wise.

As a regularization term, the predicted shared and private components are then
concatenated and fed into a reconstructor DNN $M_r$ with parameters $\theta_r$ to
recover the input speech frames $x^s$ and $x^t$ from both source and target domains
respectively. The reconstructor is trained to minimize the mean square error based
reconstruction loss as follows:
\begin{align}
	&\mathcal{L}_{\text{recon}}(\theta_c, \theta^s_p, \theta^t_p, \theta_r)
	\nonumber \\
	& \quad \quad \quad \quad = \sum_{i}^{N_s} || \hat{x}^s_i - x^s_i ||_2^2 + 
	\sum_{j}^{N_t} || \hat{x}^t_j - x^t_j ||_2^2 \\
	&\hat{x}_i^s = M_r([M_c(x^s_i), M_p^s(x^s_i)]) \\ 
	&\hat{x}_j^t = M_r([M_c(x^t_j), M_p^t(x^t_j)])
	\label{eqn:recon_loss}
\end{align}
where $[\cdot, \cdot]$ denotes concatenation of two vectors.

The total loss of DSN is formulated as follows and is jointly optimized with respect to
the parameters.
\begin{align}
	& \mathcal{L}_{\text{total}}(\theta_y, \theta_c, \theta_d, \theta_p^s,
		\theta_p^t, \theta_r)  = \mathcal{L}_{\text{senone}}(\theta_c,
		\theta_y) +
	\mathcal{L}^d_{\text{domain}}(\theta_d) \nonumber \\ 
	& - \alpha \mathcal{L}^{c}_{\text{domain}}(\theta_c)
	 + \beta \mathcal{L}_{\text{diff}}(\theta_c, \theta_p^s, \theta_p^t) 
	 + \gamma \mathcal{L}_{\text{recon}}(\theta_c, \theta_p^s, \theta_p^t,
	 \theta_r) \\
	& \min_{\theta_y, \theta_c, \theta_d, \theta_p^s, \theta_p^t, \theta_r}
		\mathcal{L}_{\text{total}}(\theta_y, \theta_c, \theta_d, \theta_p^s,
		\theta_p^t, \theta_r)
\end{align}
All the parameters of DSN are jointly optimized through backprogation with
stochastic gradient descent (SGD) as follows:
\begin{align}
	& \theta_c \leftarrow \theta_c - \mu \left[ \frac{\partial
		\mathcal{L}_{\text{senone}}}{\partial \theta_c} - \alpha \frac{\partial
			\mathcal{L}^{c}_{\text{domain}}}{\partial \theta_c} + \beta \frac{\partial
		\mathcal{L}_{\text{diff}}}{\partial \theta_c} + \gamma \frac{\partial
		\mathcal{L}_{\text{recon}}}{\partial \theta_c} \right]
		\label{eqn:grad_f} \\
	& \theta_d \leftarrow \theta_d - \mu \frac{\partial
		\mathcal{L}^d_{\text{domain}}}{\partial \theta_d}, \quad
	\theta_y \leftarrow \theta_y - \mu \frac{\partial
		\mathcal{L}_{\text{senone}}}{\partial \theta_y} \\ 
	& \theta^s_p \leftarrow \theta^s_p - \mu \left[ \beta \frac{\partial
		\mathcal{L}_{\text{diff}}}{\partial \theta^s_p}
	+ \gamma \frac{\partial \mathcal{L}_{\text{recon}}}{\partial \theta^s_p}
\right] \\
	& \theta^t_p \leftarrow \theta^t_p - \mu \left[ \beta \frac{\partial
		\mathcal{L}_{\text{diff}}}{\partial \theta^t_p}
	+ \gamma \frac{\partial \mathcal{L}_{\text{recon}}}{\partial \theta^t_p}
\right] \\
	& \theta_r \leftarrow \theta_r - \mu \frac{\partial
		\mathcal{L}_{\text{recon}}}{\partial \theta_r}
	\label{eqn:sgd}
\end{align}

Note that the negative coefficient $-\alpha$ in Eq. \eqref{eqn:grad_f} induces
reversed gradient that maximizes the domain classification loss in Eq.
\eqref{eqn:domain_loss_adv} and makes the shared components
domain-invariant.
Without the reversal gradient,  SGD would make representations
different across domains in order to minimize Eq. \eqref{eqn:domain_loss}.  For easy implementation,
GRL is introduced in \cite{grl_ganin}, which acts as an identity transform in the
forward pass and multiplies the gradient by $-\alpha$ during the backward pass.

The optimized shared component extractor $M_c$ and senone classifier $M_y$ form the
adapted acoustic model for robust speech recognition.


\section{Experiments}
\label{sec:experiment}
In this work, we perform the \emph{pure} unsupervised environment adaptation of the DNN-HMM acoustic
model with domain separation networks for robust speech recognition on CHiME-3 dataset.

\subsection{CHiME-3 Dataset}
The CHiME-3 dataset is released with the 3rd CHiME speech Separation and
Recognition Challenge \cite{chime3_barker}, which incorporates the Wall
Street Journal corpus sentences spoken in challenging
noisy environments, recorded using a 6-channel tablet based microphone
array.
CHiME-3 dataset consists of both real and simulated data. The real speech data was recorded in four real noisy
environments (on buses (BUS), in caf\'{e}s (CAF), in pedestrian areas (PED), and at street
junctions (STR)). To generate the simulated data, the clean speech is first
convoluted with the estimated impulse response of the environment and
then mixed with the background noise separately recorded in that
environment \cite{chime3_hori}. The noisy training data consists of 1600 real
noisy utterances from 4 speakers, and 7138 simulated noisy utterances
from 83 speakers in the WSJ0 SI-84 training set recorded in 4 noisy
environments. There are 3280 utterances in the development set including
410 real and 410 simulated utterances for each of the 4 environments.
There are 2640 utterances in the test set including 330 real and 330
simulated utterances for each of the 4 environments. The speakers in
training set, development set and the test set are mutually different
(i.e., 12 different speakers in the CHiME-3 dataset). The training,
development and test data sets are all recorded in 6 different channels. 

8738 clean utterances corresponding to the 8738 noisy training utterances in the CHiME-3 dataset are selected from the WSJ0 SI-85 training set to form the clean training data in our experiments. WSJ 5K word 3-gram language model is used for decoding. 


\subsection{Baseline System}
\label{sec:baseline}
In the baseline system, we first train a DNN-HMM acoustic model with clean speech and
then adapt the clean acoustic model to noisy data using GRL unsupervised adaptation
in \cite{grl_ganin}. Hence, the source domain is with clean speech while the target domain is with noisy speech.

The 29-dimensional log Mel filterbank features together with 1st and 2nd order 
delta features (totally 87-dimensional) for both the clean and noisy
utterances are extracted by following the process in \cite{li2012improving}. Each frame is spliced together
with 5 left and 5 right context frames to form a 957-dimensional feature. The spliced
features are fed as the input of the feed-forward DNN after global mean and variance
normalization. The DNN has 7 hidden layers with 2048 hidden units for each layer. The
output layer of the DNN has 3012 output units corresponding to 3012 senone labels.
Senone-level forced alignment of the clean data is generated using a GMM-HMM system.
The DNN is first trained with 8738 clean training utterances in CHiME-3 and the
alignment to minimize the cross entropy loss and then tested with simulation and real
development data of CHiME-3.

The DNN well-trained with clean data is then adapted to the 8738 noisy utterances
from Channel 5 using GRL method. No senone alignment of the
noisy adaptation data is used for the unsupervised adaptation. The feature extractor
is initialized with the first 4 hidden layers of the clean DNN and the senone
classifier is initialized with the last 3 hidden layers plus the output layers of the
clean DNN.  The domain classifier is a feedforward DNN with two hidden layers and
each hidden layer has 512 hidden units.  The output layer of the domain classifier
has 2 output units representing source and target domains. The 2048 hidden units of
the $4^{\text{th}}$ hidden layer of the DNN acoustic model is fed as the input to the
domain classifier. A GRL is inserted in between the deep representation and the
domain classifier for easy implementation.  The GRL adapted system is tested on real
and simulation noisy development data in CHiME-3 dataset.

\subsection{Domain Separation Networks for Unsupervised Adaptation}
We adapt the clean DNN acoustic model trained in Section \ref{sec:baseline} to the 8738 noisy
utterances using DSN. No senone alignment of the noisy adaptation data is used for the
unsupervised adaptation. 

The DSN is implemented with CNTK 2.0 Toolkit \cite{cntk}. The shared component
extractor $M_c$ is initialized with the first $N_h$ hidden layers of the clean DNN
and the senone classifier $M_y$ is initialized with the last $(7-N_h)$ hidden layers
plus the output layer of the clean DNN. $N_h$ indicates the position of shared
component in the DNN acoustic model and ranges from $3$ to $7$ in our 
experiments. The domain classifier $M_d$ of the DSN has exactly the same
architecture as that of the GRL.

The private component extractors $M_p^s$ and $M_p^t$ for the clean and noisy domains are
both feedforward DNNs with 3 hidden layers and each hidden layer has 512 hidden units.
The output layers of both $M_p^s$ and $M_p^t$ have 2048 output units. The
reconstructor $M_r$ is a feedforward DNN with 3 hidden layers and each hidden layer
has 512 hidden units. The output layer of the $M_r$ has 957 output units with no
non-linear activation functions to reconstruct the spliced input features. 

The activation functions for the hidden units of $M_c$ is sigmoid. The activation
functions for hidden units of  $M^s_p$, $M^t_p$, $M_d$ and $M_r$ are rectified linear
units (ReLU).  The activation functions for the output units of $M_c$ and $M_d$ are
softmax. The activation functions for the output units of $M^s_p$, $M^t_p$ are
sigmoid. All the sub-networks except for $M_y$ and $M_c$ are randomly initialized.
The learning rate is fixed at $5 \times 10^{-5}$ throughout the experiments.  The adapted DSN
is tested on real and simulation development data in CHiME-3 Dataset.
\begin{table}[h]
\centering
\begin{tabular}[c]{c|c|c|c|c|c|c}
	\hline
	\hline
	System & Data & BUS & CAF & PED & STR & Avg.\\
	\hline
	\multirow{2}{*}{\begin{tabular}{@{}c@{}} Clean
		\end{tabular}} & Real & 36.25 & 31.78 & 22.76 & 27.18 & 29.44  \\
	\hhline{~------}
	& Simu & 26.89 & 37.74 & 24.38 & 26.76 & 28.94 \\
	\hline
	 \multirow{2}{*}{\begin{tabular}{@{}c@{}} GRL 
		\end{tabular}} & Real & 35.93 & 28.24 & 19.58 & 25.16 & 27.16 \\ 
	\hhline{~------}
	& Simu & 26.14 & 34.68 & 22.01 & 25.83 & 27.16 \\
	\hline
	\multirow{2}{*}{\begin{tabular}{@{}c@{}} DSN
	\end{tabular}} & Real & 32.62 & 23.48 & 17.29 & 23.46 & \textbf{24.15} \\
	\hhline{~------}
	& Simu & 23.38 & 30.39 & 19.51 & 22.01 & \textbf{23.82} \\
	\hline
	\hline
\end{tabular}
  \caption{The WER (\%) performance of unadapted acoustic model, GRL and DSN adapted
	  DNN acoustic models for robust ASR on real and simulated development set of CHiME-3.}
\label{table:asr_wer}
\end{table}

\begin{table*}[!t]
\centering
\begin{tabular}[c]{c|c|c|c|c|c|c|c|c|c|c}
	\hline
	\hline
        \multirow{2}{*}{\begin{tabular}{@{}c@{}} $N_h$ 
	\end{tabular}} & \multicolumn{10}{c}{Reversal Gradient Coefficient $\alpha$} \\
	\hhline{~----------}
          & 1.0& 2.0& 3.0& 4.0& 5.0& 6.0& 7.0& 8.0& 9.0& Avg. \\
	\hline
        3& 27.2& 26.24& 25.76& 26.51& 26.12& 26.92& 26.65& 26.91& 27.41& 26.64 \\
	\hline
        4& 26.56& 26.08& 25.75& 25.99& 25.88& 26.76& 27.0& 27.13& 27.74& 26.54 \\
	\hline
        5& 26.53& 25.9& 26.07& 25.88& 25.72& 26.17& 27.36& 26.67& 27.37& 26.41 \\
	\hline
        6& 25.77& 25.17& 25.06& 24.94& 24.6& 25.19& 25.53& 25.42& 25.93& 25.29 \\
	\hline
	7& 25.99& 25.5& 24.73& 24.43& 25.08& 24.53& 25.07& 24.15& 24.29& \textbf{24.86} \\
	\hline
\end{tabular}
  \caption{The ASR WERs (\%) for the DSN adapted acoustic models with respect to
	  $N_h$ reversal gradient coefficient $\alpha$ on the real development set of
	  CHiME-3.}
\label{table:wer_nh_alpha}
\end{table*}

\subsection{Result Analysis}

Table \ref{table:asr_wer} shows the WER performance of clean, GRL adapted and
DSN adapted DNN acoustic models for ASR. The clean DNN achieves 29.44\% and 28.25\% WERs on
the real and simulated development data respectively. The GRL adapted acoustic model
achieves 27.16\% and 27.16\% WERs on the real and simulated development data. The
best WER performance for DSN adapted acoustic model are 24.15\% and 23.82\% on
real and simulated development data, which achieve 11.08\% and 12.30\% relative
improvement over the GRL baseline system and achieve 17.97 \% and 17.69\% relative
improvement over the unadapted acoustic model. The best WERs are achieved when
$N_h = 7$ and $\alpha = 8.0$. By comparing the GRL and DSN performance at $N_h = 4$, we observe that the introduction of private components and reconstructor lead to 5.1\% relative improvements in WER.


We investigate the impact of shared component position $N_h$ and the reversal
gradient coefficient $\alpha$ on the WER performance as in Table
\ref{table:wer_nh_alpha}.  We observe that the WER decreases with the growth of $N_h$, which is
reasonable as the higher hidden representation of a well-trained DNN acoustic model is
inherently more senone-discriminative and domain-invariant than the lower layers and
can serve as a better initialization for the DSN unsupervised adaptation.



\section{Conclusions}
In this work, we investigate the domain adaptation of the DNN acoustic model by using domain separation networks.  Different from the conventional supervised, semi-supervised and T/S adaptation approaches, DSN is capable of adapting the acoustic model to the adaptation data without any exposure to its transcription, decoded lattices or unlabeled parallel data from the source domain. The shared component between source and
target domains extracted by DSN through adversarial multi-task training is both
domain-invariant and senone-discriminative. The extraction of private component that
is unique to each domain significantly improves the degree of domain-invariance and
the ASR performance.

When evaluated on the CHiME-3 dataset for environment adaption task, the DSN achieves 11.08\% and 17.97\% relative WER improvement over the GRL baseline system and the unadapted acoustic model. The WER decreases when higher hidden representations of the DNN acoustic model are used as the initial shared component. The WER first decreases and then increases with the growth of the reversal gradient coefficient.

In the future, we will adapt long short-term memory-recurrent neural networks acoustic models \cite{sak2014long, meng2017deep, erdogan2016multi} using DSN and compare the improvement with the feedforward DNN. Moreover, we will perform DSN-based unsupervised adaptation with thousands of hours of data to verify its scalability to large dataset.

\vfill\pagebreak
\clearpage

\bibliographystyle{IEEEbib}
\bibliography{strings,refs}

\end{document}